\documentclass{article}

\usepackage{hyperref}

\usepackage[accepted]{icml2025}

\usepackage{graphicx}
\usepackage{array}
\usepackage{float}
\usepackage{multirow}
\usepackage{subcaption}
\usepackage{amsthm}
\usepackage{amssymb}

\usepackage{wrapfig}
\usepackage{mathtools}

\usepackage[capitalize,noabbrev]{cleveref}

\icmltitlerunning{Towards Personalized Treatment Plan: Geometrical Model-Agnostic Approach to Counterfactual Explanations}

\begin{document}

\twocolumn[
\icmltitle{Towards Personalized Treatment Plan: Geometrical Model-Agnostic Approach to Counterfactual Explanations}



\icmlsetsymbol{equal}{*}

\begin{icmlauthorlist}
\icmlauthor{Daniel Sin}{author}
\icmlauthor{Milad Toutounchian}{investigator}
\end{icmlauthorlist}

\icmlaffiliation{author}{B.S. (Drexel University) \& Computer and Information Science (CIS) Master's Student at University of Pennsylvania, Philadelphia, United States}
\icmlaffiliation{investigator}{Principal Investigator, Associate Professor at Drexel University, Philadelphia, United States}

\icmlcorrespondingauthor{Daniel Sin}{sin24@seas.upenn.edu}
\icmlcorrespondingauthor{Milad Toutounchian}{mt3393@drexel.edu}

\icmlkeywords{Machine Learning, ICML}

\vskip 0.3in
]



\printAffiliationsAndNotice{}  

\begin{abstract}
What if a person never smoked? Would they have lung cancer? What identifies those who don't have cancer from those who don't? Counterfactuals are statements of ``alternative hypothesis" that describe the outcomes of an alternative scenario \cite{pearl2018book}. They question the possibilities of a scenario that never occurred. While counterfactuals are hard to analyze, we can analyze counterfactual explanations which are more rigorously based. Counterfactual explanations define the ``smallest" change to a data point such that the change flips the predicted outcome for a data point \cite{molnar2025}. In our article, we describe a method for generating counterfactual explanations in high-dimensional spaces using four steps that involve fitting our dataset to a model, finding the decision boundary, determining constraints on the problem, and computing the closest point (counterfactual explanation) from that boundary. We propose a discretized approach where we find many discrete points on the boundary and then identify the closest feasible counterfactual explanation. This method, which we later call \textit{Segmented Sampling for Boundary Approximation} (SSBA), applies binary search to find decision boundary points, filters the decision boundary points for feasible points, and then returns the closest feasible point. Across four datasets of varying dimensionality, we show that our method can outperform current methods for counterfactual generation with reductions in distance between $5\%$ to $50\%$ in terms of the $L_2$ norm. Our method can also handle real-world constraints by restricting changes to immutable and categorical features, such as age, gender, sex, height, and other related characteristics such as the case for a health-based dataset. In terms of runtime, the SSBA algorithm generates decision boundary points on multiple orders of magnitude in the same given time when we compare to a grid-based approach. In general, our method provides a simple and effective model-agnostic method that can compute nearest feasible (i.e. realistic with constraints) counterfactual explanations. All of our results and our code can be found here at this link: \href{https://github.com/dsin85691/SSBA_For_Counterfactuals}{https://github.com/
dsin85691/SSBA\_For\_Counterfactuals}
\end{abstract}

\section{Introduction}

Imagine two scenarios: ``If Joe had not smoked for $30$ years, he would have lived," or ``if Joe were 30 pounds lighter, he would not have heart disease." While we may know that Joe smoked or was 30 pounds heavier, we may ask a question where we imagine outcomes to scenarios where the initial conditions were altered. These scenarios are among many described by Dr. Judea Pearl as ``counterfactuals." Counterfactuals are statements of ``alternative hypothesis" that question the outcome given an alternative set of initial conditions \cite{pearl2018book}. When compared to other types of causal relationships commonly studied in machine learning, such as association or intervention, counterfactuals are ``problematic" for study since we deal with outcomes that are not easily identifiable from the dataset. When given observational data for a person, we can never observe more than one potential outcome for a given person \cite{pearl2018book}. This makes it challenging to create logically consistent counterfactuals in a setting where we cannot perfectly replicate the same inputs. Standard machine learning methods, such as regression, can be helpful for questions of association where we observe how, for a set of variables $X$ and $Y$, a change in $X$ can alter the value of $Y$. With intervention, it is common to set up conditions so that other variables do not affect the primary variable of interest and its corresponding dependent variable. Randomized Control Trials (RCTs) were considered the ``gold standard of causal analysis," and it was common to randomize subjects into different groups to minimize the effect of confounding variables on treatment \cite{pearl2018book}. However, with counterfactuals, we must estimate the relationships between variables in a world where ``observed facts are bluntly negated" \cite{pearl2018book}. 

To compute counterfactuals, Dr. Pearl relied on a method that made the use of structural causal models (SCMs) which are graphical models consisting of exogenous variables ($U$) and endogenous variables ($V$) and a ``set of functions $f$ that assigns each variable in $V$ a value based on the other variables in the model" \cite{pearl2016causal}. Dr. Pearl provides a three-step process for computing any deterministic counterfactual with a structural causal model $M$, which includes abduction, action, and prediction \cite{pearl2016causal}. Abduction uses evidence $E = e$ to set the values of the exogenous variables in $M$. Action modifies the original model $M$ by removing the structural equations for the variables in $X$ and replacing them with the appropriate functions $X = x$ to obtain the model $M_x$ \cite{pearl2016causal}. Prediction uses the modified $M_x$ and the value of exogenous variables in $U$ to compute the value of $Y$, the consequence of the counterfactual \cite{pearl2016causal}. Similar to the idea used for Bayesian networks, we can use evidence or conditional information on the exogenous variables to simplify and construct a modified model that we can use to predict the outcome. 

In high-dimensional datasets, determining the structural causal model beforehand can be challenging. The use of causal discovery algorithms may require conditions such as causal sufficiency, which necessitates knowing all confounding variables within the dataset. Due to the difficulty of building a structural causal model and inferring causal relationships directly from the dataset, we address these issues by simplifying the problem to binary classification, assuming sufficient data for both classes, and relying on a formal definition of counterfactual explanations. As described by Dandl and Molnar [2025], a counterfactual explanation (CFE) describes ``the smallest change to the feature values that changes the prediction to a predefined output" \cite{molnar2025}. Given the direct causal relationship between input and prediction of a machine learning model, we can observe ``scenarios in which the prediction changes in a relevant way, like a flip in predicted class" \cite{molnar2025}. For CFEs, we find the smallest change across features such that the model flips a prediction value for a specific data point. For binary classification, such a flip would mean becoming a member of the other class as predicted by the model. In terms of criteria, we would like these explanations to be ``as similar as possible to the instance regarding feature values" and require us to change as ``few features as possible" \cite{molnar2025}. Finally, we would want feature values that are likely for the counterfactuals \cite{molnar2025}. Producing a CFE that generates illogical features, such as a younger person, or a person with an age greater than $150$, would not be realistic. 

To generate a CFE, we can employ multiple approaches, including randomly selecting counterfactuals, optimizing the choice of counterfactuals, and generating decision boundary points (our approach). One approach involves ``randomly changing feature values of the instance of interest and stopping when the desired output is predicted" \cite{molnar2025}. As another approach, Wachter et. al. [2017] proposed an objective loss function where the two terms of the objective reduce the quadratic distance between the model prediction for the counterfactual $\mathbf{x'}$ and a second term to minimize the distance between the instance $\mathbf{x}$ and the counterfactual $\mathbf{x'}$ \cite{molnar2025, wachter2017counterfactual}. Across multiple iterations, we would find a counterfactual $x'$ that minimizes the objective loss function, and then we return that counterfactual at the end. \\
\begin{align*}
L(\mathbf{x}, \mathbf{x}^{\prime}, y^{\prime}, \lambda) = \lambda \cdot (\hat{f}(\mathbf{x}^{\prime}) - y^{\prime})^2 + d(\mathbf{x}, \mathbf{x}^{\prime}) \\
\end{align*}
Dandl et. al. [2020] also implemented a four-objective loss function that minimized all of the above criteria for counterfactual explanations \cite{molnar2025}. Similar to Wachter et. al. [2017], they applied an combined objective loss function to optimize an appropriate counterfactual $\mathbf{x'}$ selectively.  $\mathbf{o_1}, \cdots \mathbf{o_4}$ are objective functions that each individually represents one of the criteria above for CFEs. \\
\begin{align*}
L(\mathbf{x},\mathbf{x'},y',\mathbf{X}^{\text{obs}})=\big(o_1(\hat{f}(\mathbf{x'}),y'),o_2(\mathbf{x}, \mathbf{x'}),o_3(\mathbf{x},\mathbf{x'}), \\
o_4(\mathbf{x'},\mathbf{X}^{\text{obs}})\big) \\
\end{align*}
Among model-agnostic methods developed, Mothilal, Sharma, and Tan implemented DiCE (Diverse Counterfactual Explanations for ML), a Python package that generates CFEs for both gradient-based and model-agnostic methods \cite{mothilal2020dice}. Their gradient-based method optimizes over a combined loss function over all generated counterfactuals that take into account the difference between the machine predictions $f(c_i)$ and the desired outcomes $y$ (first term), the distance between the counterfactual $c_i$ and the instance $x$, and the diversity among different counterfactuals. 

\begin{align*}
C(x) = \arg\min_{c_1, \ldots, c_k}
    & \frac{1}{k} \sum_{i=1}^{k} \text{yloss}(f(c_i), y) \\
    + \lambda_1 \cdot \frac{1}{k} \sum_{i=1}^{k} \text{dist}(c_i, x)
    & - \lambda_2 \cdot \text{dpp\_diversity}(c_1, \ldots, c_k)
\end{align*}\cite{mothilal2020dice}

In their model-agnostic approaches, they adopt three different algorithms to generate a multitude of CFEs, which are their KDTrees algorithm, their random algorithm, and their genetic algorithm. The KDTree algorithm queries the $k$ closest counterfactuals from the dataset using KDTrees \cite{mothilal2020dice}. The random algorithm generates a random set of counterfactuals that are generally in the opposite class relative to the original instance \cite{mothilal2020dice}. Finally, the genetic algorithm tries to find the best counterfactuals close to the query point using a convergent algorithm that computes the best counterfactuals for a given generation based on a loss function taking into account criteria such as proximity and sparsity. It then generates a new generation based on $50\%$ of the fittest members of the current generation \cite{mothilal2020dice}. All of these methods use the original instance $\mathbf{x}$ and the counterfactual $\mathbf{x'}$ to iterate and find a new counterfactual that matches the criteria and reduce the objective loss function between the counterfactual and the original instance. 

While other methods focus on optimization of a counterfactual $\mathbf{x'}$, we propose a method that focuses first on finding decision boundary points and then computing the counterfactual. We find such points by applying binary search on pairs of points from differing classes. By doing so, we can find approximate, discrete points on the decision boundary that are between two points of differing predicted classes within a sufficiently small $\epsilon$. A sufficiently small $\epsilon$ indicates that there exists a decision boundary point that separates these points. Once we have iterated over a sufficient number of points with differing target values, we can compute the closest counterfactual explanation using $L_2$ norm. Across all decision boundary points, we find the $d^* \in D$ such that $min_{d \in D} \ dist(x, d)$ is minimized, where $x$ is the instance for which we compute the counterfactual explanation. Once we have that decision boundary point $d^*$, we compute the closest CFE by taking an epsilon $\epsilon_0$ sufficiently small such that $f(d^* + \epsilon_0) \neq y$ where $y$ is the class for $x$.\\

\section{Related Works}

In this article, we primarily examine our methods with respect to DiCE and Alibi, which are both open-source Python packages for high-dimensional counterfactual generation. Wexler et al. [2019] present a similar methodology for nearest counterfactual generation. However, their approach selects a data point of a differing target class within the dataset \cite{wexler2019if}. While their tool can generate nearest counterfactual explanations based on the dataset, their ``What-If" tool is specialized for two dimensions and does not readily apply to higher dimensions. Karimi et. al. [2019] implemented MACE (model-agnostic counterfactual explanation) with the use of SMT solvers, but their work applies a distance metric based on $L_0$, $L_1$, and $L_{\infty}$ and not $L_2$ for $n \geq 2$ \cite{karimi2020model}. Additionally, the discontinuation of the MACE project has made it difficult to replicate results applying $L_2$ norms. Due to limited research on counterfactual generation using $L_2$, we limit our work with respect to DiCE and Alibi.

\section{Methodology}

Our methodology is based on a discretized approach to find the nearest feasible counterfactual explanation.

We explore two different appproaches to generating decision boundary points: one where we generate a grid $G \subset \mathbf{R}^n$ and where $|G| = R^n$ where $R$ is the size of a single dimension and $n$ is the number of features and one where we use binary search to compute decision boundary points from pairs of points from differing classes. We use a small $\epsilon$, usually $1.0 * 10^{-3}$, so that the chosen decision boundary point is within a small distance between two points of different classes. After generating a discrete set of decision boundary points, we then apply the KD Trees algorithm from the sci-kit learn library to compute the closest decision boundary point from the original instance. 

In the results section, we compare the computed counterfactual explanations and the runtime for various counterfactual generation methods which include DiCE, Alibi, and our SSBA method. As discussed earlier, Mothital et al. released DiCE, an open-source repository that adopts much of the recent research in the space of generating CFEs. Additionally, Janis Klaise et. al. [2021] created Alibi, an open-source library for the generation of CFEs, and provided an open-source implementation of the loss function from Wachter et. al. [2017] \cite{JMLR:v22:21-0017}. They apply gradient descent to the objective loss function for correcting the distance between the initial counterfactual $\mathbf{x'}$ and the original instance $\mathbf{x}$ \cite{JMLR:v22:21-0017}. 

\begin{figure}[H]
    \centering
    \includegraphics[width=3.25in, height=2in]{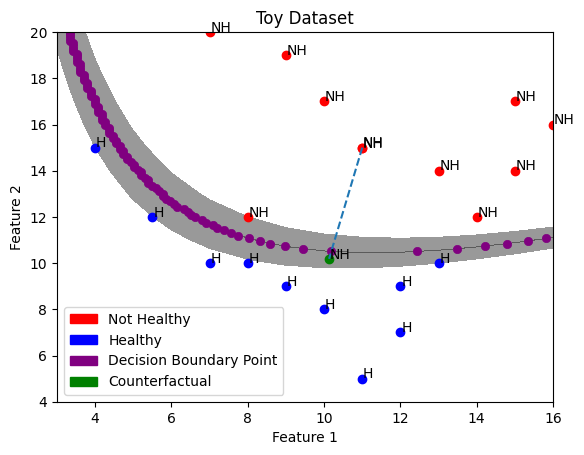}
    \caption{Counterfactual explanation for exemplary point $(11,15)$ crossing a trained SVM's decision boundary. We generate the points in purple to find the closest point to the boundary.}
\end{figure}

\subsection{Handling categorical features}

For training the models, we treat the categorical features as integers where each number is representative of a category for a specific feature. For computing the decision boundary points using a trained model, we leave the categorical features as floating-point numbers and avoid rounding our boundary point. For searching through the set of decision boundary points, we usually remove any points that do not match the categorical features of the original query instance. If we cannot find any points matching the values of immutable features, we restrict all categorical features from changing and modify only the continuous features. Instead of moving the full displacement to the boundary, we generate a box around the original point, namely a set of points $\Delta \mathbf{x} \subset \mathbf{R}^n$, and we find the closest point to the boundary from this given box. We define $\Delta x_i$ for each $i \in \{1, \cdots, n\}$ such that $\Delta x_i$ is some percentage of the original point. In the case of the paper, we search through points $(x - \Delta x, x + \Delta x)$ where ($x_i - \Delta x_i$, $x_i + \Delta x_i$) $\forall i \in {1, \cdots, n}$ and where $\Delta x_i = 0.2x_i$. 

\subsection{Segmented Sampling for Boundary Approximation Algorithm}
 
\begin{algorithm}[H]
\renewcommand{\thealgorithm}{}
\caption{Bisection Method Algorithm: Method of Boundary Point Generation}
\begin{algorithmic}[1]
\REQUIRE Trained model $f$, dataset $X$, labels $y$, threshold $T$, precision $\epsilon$, $n$ the number of features
\ENSURE Set of approximated boundary points
\STATE Identify class labels $c_1$, $c_2$ from $y$ \\
\STATE Split $X$ into $X_{c_1}$ and $X_{c_2}$ according to $y$
\STATE Select correctly predicted points from each class: 
    \begin{itemize}
        \item $X_{c_1}^{\text{correct}} = \{\mathbf{x} \in X_{c_1} \subset \mathbf{R}^n \mid f(\mathbf{x}) = c_1 \}$
        \item $X_{c_2}^{\text{correct}} = \{\mathbf{x} \in X_{c_2} \subset \mathbf{R}^n \mid f(\mathbf{x}) = c_2 \}$
    \end{itemize}
\STATE Randomly sample $N$ pairs $(\mathbf{x}_a, \mathbf{x}_b) \subset X_{c_1}^{\text{correct}} \times X_{c_2}^{\text{correct}}$ \\
\STATE $D \gets \text{empty list}$ 
\FOR{$(\mathbf{x_a}, \mathbf{x_b}) \subset X_{c_1}^{\text{correct}} \times X_{c_2}^{\text{correct}}$}
    \STATE Compute $\alpha \gets$ \newline $\text{AlphaBinarySearch}(f, \mathbf{x}_a, \mathbf{x}_b, c_1, \epsilon)$
    \STATE Compute boundary point: $\mathbf{x}_\text{boundary} = (1 - \alpha)\mathbf{x}_a + \alpha \mathbf{x}_b$
    \STATE Append $\mathbf{x}_\text{boundary}$ to $D$
\ENDFOR \\
\STATE Return $D$, Discrete list of boundary points
\end{algorithmic}
\end{algorithm}

\begin{algorithm}[bt]
\renewcommand{\thealgorithm}{}
\caption{Alpha Binary Search: Binary Search on Model Prediction}
\begin{algorithmic}[1]
\REQUIRE Trained binary classifier $f$, points $\mathbf{x}_a$, $\mathbf{x}_b$, label $y_a$, tolerance $\epsilon$, maximum iterations $T$
\ENSURE $\alpha \in [0, 1]$ such that $|r-l| < \epsilon$
\STATE $l \gets 0,\ r \gets 1$
\FOR{$t = 1$ \text{ to } $T$}
    \STATE $\alpha \gets \frac{l + r}{2}$
    \STATE $\mathbf{x}_\alpha \gets (1 - \alpha)\mathbf{x}_a + \alpha \mathbf{x}_b$
    \STATE $\hat{y} \gets f(\mathbf{x}_\alpha)$
    \IF{$\hat{y} = y_a$}
        \STATE $l \gets \alpha$
    \ELSE
        \STATE $r \gets \alpha$
    \ENDIF
    \IF{$|r - l| < \epsilon$}
        \STATE \textbf{break}
    \ENDIF
\ENDFOR
\STATE Return $\alpha$
\end{algorithmic}
\end{algorithm}
\newpage
\subsubsection{Algorithm Overview}

In this section, we first apply the binary search method (bisection method) as suggested by Vlassopoulous et. al. \cite{DBLP:journals/corr/abs-2006-07985}. This generates a discrete set of decision boundary points for a binary classification problem. Our variant of the algorithm takes in a trained model (model-agnostic), a dataset, labels, a threshold (number of total boundary points), and a precision value for the $\alpha$ used in the algorithm. 

It requires splitting the dataset into two discrete sets, consisting of those labelled as the first class $c_1$ and those of the other class $c_2$. We separate all such points into two different classes $X_{c_1}^{\text{correct}} \subset \mathbf{R}^n$ and  $X_{c_2}^{\text{correct}} \subset \mathbf{R}^n$ such that they are both correctly predicted. We then sample pairs of points ($\mathbf{x}_a$, $\mathbf{x}_b$) from differing classes. Equivalently, we sample a set of points of the form ($\mathbf{x}_a$, $\mathbf{x}_b$) from $X_{c_1}^{\text{correct}} \times X_{c_2}^{\text{correct}}$. We then compute a value $\alpha$ that exists on a line segment between $\mathbf{x}_a$ and $\mathbf{x}_b$, which we then use to compute a decision boundary point on line $7$. 

The second algorithm applies binary search such that the endpoints $l$ and $r$ are from differing classes. At the end of the algorithm, we return an $\alpha$ that represents the location of a decision boundary point on the line segment. We then use $\alpha$ on line $8$ to determine adding a new boundary point.

\subsection{Ensuring feasibility compatible CF explanation method} 

Mahajan et. al. [2019] provide a method whereby they define additional example loss terms to their optimization. For unary constraints (single term constraints), they add a hinge loss $-min (0, x_v^{'} - x_v)$ where $v$ selects a specific feature and $'$ superscript refers to the counterfactual of the instance $x$. For binary constraints, they use a linear model to measure the causal relationship between two features \cite{mahajan2019preserving}.  

Mothilal, Sharma, and Tan [2020] allowed for two types of constraints in their implementation of DiCE. One constraint type allowed setting ``feasible ranges for each [continuous] feature." Another constraint type allows users to choose which variables or features can be changed while leaving others unchanged \cite{mothilal2020dice}. 

We let users define constraints in two different ways. In the first case, we allow users to define equality, less than, and greater than constraints for real-world constraints that constrain the choice of CFEs to fit reality. Mahajan et al. [2019] provide examples of unary constraints, such as $x_{age}^{cf} \geq x_{age}$, which restrict the choice of CFEs to either have the same age or older when compared to the age of the original instance \cite{mahajan2019preserving}. 

In the second case, we also apply the ``box constraints" for continuous features that DiCE applies. However, when we define these constraints, we define a set of delta constraints around the instance $x = (x_1, \cdots, x_n)$ such that for delta constraints $\Delta x_i$, ($x_i - \Delta x_i$, $x_i + \Delta x_i$) is the range of features that we allow for. 

In the case of finding the nearest feasible counterfactual explanation, we remove decision boundary points $d \in D$ from our search space that do not match our constraints. Rather than applying a loss function, we use the constraints to reduce the number of discrete points on the decision boundary that we would have to check for. We share examples of these constraints and their usage in the results section. 

\section{Results}

\subsection{Datasets}

For the datasets used in Table $1(a)$, we used sci-kit's ``make\_classification" function to generate three synthetic datasets of varying dimensionality consisting of $2$, $10$, and $50$ features. We used these synthetic datasets to compare the runtimes of different model-agnostic approaches, including the previously discussed grid-based approach and DiCE's model-agnostic approaches.

For Table $1(b)$, we make use of four different datasets. The first dataset is the one dataset that we created, as shown in Figure $1$, which consists of $20$ data points belonging to two classes, ``unhealthy" and ``healthy." The ``healthy" data points are colored blue, and the ``unhealthy" data points are colored red as shown in Figure $1$. This dataset is heavily used for visualization purposes, and we use this dataset to see visual differences among methods in the two-dimensional case. 

Secondly, the Adult Income dataset comprises over 26,000 fully labelled data points, each with an associated list of observational features and target values indicating whether the person's annual income exceeds 50,000. We use DiCE's variation of the dataset, which consists of $8$ features and one target value \cite{adult_2}. 

Thirdly, we use the heart disease dataset that consists of $13$ characteristics and one target value that indicates the presence or absence of heart disease. This dataset comprises over $500$ data points, with $13$ features representing various characteristics of the person, including age, sex, type of chest pain (cp), serum cholesterol levels, and maximum heart rate (thalach), among others \cite{heart_disease_45}. 

Finally, we generated a synthetic dataset of $20$ continuous features of two different classes, namely class labels $0$ and $1$, with sci-kit learn. For the experiments, we used $2000$ data points, with each class represented equally by $1000$ data points. 

The first and last datasets were used for comparing our SSBA method with DiCE and Alibi's methodologies in the case where the features were unconstrained. Given that all of the features of these two datasets are continuous features, we can easily apply each method and observe the distances of the generated counterfactuals from the original instance. With the second and third datasets, we can restrict all methods from changing any of the categorical features and only let continuous features change. 

\subsection{Comparing Runtime of Different Approaches}

In Table $1(a)$, we can see apparent differences in the results produced by the SSBA method and the grid-based approach (generating a grid $G \subset R^n$) for a varying number of features when applied to these synthetic datasets. As shown in Table $1(a)$, the grid-based approach produces memory errors for dimensions higher than two. The grid-based approach generates a grid $G \subset R^n$ where $n$ is the number of features, and in high-dimensional space, generating such a grid becomes infeasible since the memory complexity is exponential in the number of features. Additionally, the time complexity is also exponential in the number of features since the method traverses the entire grid to find points that differ in class targets. The SSBA method described in Table $1(b)$, applies binary search to find decision boundary points and does so with a much lower time and space complexity. SSBA is bounded by $O(N^2 * T_{predict}*log (\lfloor \frac{D}{\epsilon} \rfloor)))$ where $N^2$ comes from the number of possible pairs of two classes with different target values\footnote{Classes can be equally sized. For a given dataset of size $N$, it could be divided such that each class has $\frac{N}{2}$ members, and thus there are $\frac{N}{2}*\frac{N}{2} = \frac{N^2}{4}$ possible number of pairs.} , $\lfloor log ({\frac{D}{\epsilon}}) \rfloor$ comes from the number of possible iterations for binary search where $D$ represents the maximal distance between two points in the dataset, and $T_{predict}$ is the time for a single point prediction from a given machine learning model. The number of point predictions for the grid-based approach would be $O(T_{predict}*R^n)$ since we apply the model to every single point in the grid of size $R^n$.

Additionally, when applying GPUs, we also run over large batches of points at once, usually a batch of $1000$ points in the experiments. In this case, the method shows much more comparable runtimes to DiCE's methods. As shown in Table $1(a)$, the runtime columns showcase that our methods can generally match the runtime of DiCE's model-agnostic approach.

\subsection{Table Results} 

We compare our methods in two cases, unconstrained and constrained. For the unconstrained case, we compare the $L_2$ distance between the generated counterfactual and the original query instance. For the constrained case, we compare the bounded counterfactual (partial counterfactual) with the closest boundary point. This means that the generated counterfactual has features $\mathbf{x_i}$ such that for every $i$, ($x_i - \Delta x_i$, $x_i + \Delta x_i$). For multiple points, such as shown in the Table $1(b)$, we take an average of the distances produced between the generated counterfactuals and the  query points. The $L_2$ norm is represented as a scalar value that we compare for each respective dataset, and we use the four datasets discussed previously. For demonstration of its effect as a model-agnostic approach, we use four different types of models, which are trained prior to counterfactual generation, namely, support vector machines (SVMs), multi-layer perceptrons (MLPs or just neural networks), logistic regression (LR), and random forest classifiers. We computed $50,000$ boundary points for each run in the case of DiCE, whereas Alibi was limited to $25,000$ points.  

In general, our method outperforms DiCE's model-agnostic approaches for both constrained and unconstrained cases, whereas our method outperforms Alibi for constrained cases. For the unconstrained case, we compare the methods for the first and last datasets which consist of numerical, continuous values. In contrast, we compare the methods using the second and third datasets which consist of both numerical and categorical values.

In the unconstrained case, we observe that our method generally outperforms all of DiCE's model-agnostic approaches, and the method achieves smaller $L_2$ norms for the first and last datasets, which consist of $2$ and $20$ continuous features respectively, as scalar values. Our method produced similar results to Alibi for two dimensions, whereas our method does not perform as well for high dimensions (such as the fourth dataset). 

In the constrained case, our method outperforms both DiCE's and Alibi's methods for the Adult Income and Heart Disease datasets. As shown in the last column of Table $1(b)$, we find that our method produces closer bounded counterfactuals to the decision boundary. With a bounded counterfactual, we restrict the generated counterfactual $x' \in (x - \Delta x, x + \Delta x)$. We select $\mathbf{x'}$ such that the distance is minimized between $\mathbf{x'}$ and the boundary points to the decision boundary. We then take the sum of these distances and average them which are reported on the last column of Table $1(b)$.

Regarding the metrics and how we computed the last column of Table $1(b)$, we provided an appendix section that goes more into depth on how we evaluated the methods and how we computed the distances used in Table $1(b)$.

\subsection{Visual Examples} 

In figures $3$ and $4$, we share visual examples of the unconstrained case. Specifically, we compare how our method performs for the point $(11,15)$ when compared to DiCE's model-agnostic methods and Alibi's gradient methods. When compared to DiCE's model-agnostic methods, our method produces a counterfactual explanation closer to the decision boundary, and the use of discrete decision boundary points allow for us to appropriately choose one that is closest from the original instance. In figure $4$, we find that our method performs on par with Alibi's gradient-based approach. We also show the decision boundary points generated in all figures, highlighted as purple dots in figures $2$, $3$, and $4$. 

In figure $5$, we share visual examples of the constrained case. Specifically, we share some visual examples of how our method performs when we apply constraints in both $x_1$ and $x_2$ directions. For producing the bounded counterfactual as described earlier, we take a ``box" around the original data point, which means that we look at points between $(x_1 - \Delta x_1, x_2 - \Delta x_2)$ and $(x_1 + \Delta x_1, x_2 + \Delta x_2)$. We take the closest point from that ``box" to the decision boundary and compute the closest distance from that point to the boundary. We can then average this for a number of samples from our dataset. We provide further explanation of this method in the appendix where we describe the evaluation metrics.

\begin{table*}
\centering
\begin{minipage}{1\linewidth}
\centering
\begin{tabular}{|c|c|c|c|c|c|}
\hline
\textbf{Features} & \textbf{Method} & \textbf{Limit for Boundary Points} & \textbf{Runtime} & \textbf{Memory / Error} & \textbf{Number} \\
\hline
\multirow{6}{*}{50} 
  & SSBA (GPU)  & $T=1{,}000{,}000$   & ~6.2s   & -         & 689{,}574 \\
  &            & $ T=100{,}000$   & ~0.9s    & -         & 100{,}000 \\
  &            & $T=10{,}000$     & ~0.1s     & -         & 10{,}000 \\
  &   Grid-based         & $T=10^{50}$ ($R = 10$, $10^{50}$ points) & -        & Memory Error & 0 \\
& DiCE (model-agnostic) & - & 13.2s  & -         & - \\
\hline
\multirow{5}{*}{10} 
  & SSBA (GPU)  & $T=1{,}000{,}000$   & ~2.2s   & -         & 711{,}436 \\
  &            & $T=100{,}000$   & ~0.4s    & -         & 100{,}000 \\
  &            & $ T=10{,}000$     & ~0.1s     & -         & 10{,}000 \\
  &   Grid-based         & $T=10^{10}$ ($10^{10}$ total points)          & -        & 745 GB Memory Error & 0 \\
   & DiCE (model-agnostic) & - & 5.08s  & -         & - \\
\hline
\multirow{7}{*}{2}  
  & SSBA (GPU)  & $T=1{,}000{,}000$   & ~2.4s   & -         & 793{,}845 \\
  &            & $T=100{,}000$   & ~0.3s    & -         & 100{,}000 \\
  &            & $T=10{,}000$     & ~0.0s     & -         & 10{,}000 \\
  & Grid-based & $T=22,500$ ($150^2 = 22{,}500$ points) & 28.1s    & -         & 311 \\
  &  Grid-based     & $T=10{,}000$ ($100^2 = 10{,}000$ points) & 5.5s     & -         & 
  104 \\
  & DiCE (model-agnostic) & - & 0.0303s    & -         & - \\
\hline
\end{tabular}
\newline
\newline
\end{minipage}%
\hfill
\begin{minipage}{1.0\linewidth}
\centering
\begin{tabular}{|c|c|c|c|c|c|c|}
\hline
\multicolumn{7}{|c|}{Euclidean Distance for Counterfactual Explanations} \\
\hline
Algorithm & Dataset & Model & Number of Samples & Constraints & Number of CFs & Distance \\
\hline
\multicolumn{7}{|c|}{Toy Dataset} \\
\hline
SSBA & Toy & SVM & All $20$ points & No & 1 (for each point) & \textbf{3.299} \\
DiCE (kdtree) & Toy & SVM &  All $20$ points & No & 100 & 9.599 \\
DiCE (random) & Toy & SVM &  All $20$ points & No & 100 & 8.307 \\
DiCE (genetic) & Toy & SVM &  All $20$ points & No & 100 & 4.966 \\
SSBA & Toy & MLP (2 layers)  & All $20$ points & No & 1 & 1.532 \\
Alibi (Wachter Loss) & Toy & MLP (2 layers) & All $20$ points & No & 1 & \textbf{1.513} \\
\hline
\multicolumn{7}{|c|}{Adult Income Dataset} \\
\hline
SSBA &  Adult Income & Random Forest & $500$ points & Yes & 1 & \textbf{2.102} \\
DiCE (kdtree) & Adult Income & Random Forest & $500$ points   & Yes & 10  & 2.407 \\
DiCE (random) & Adult Income & Random Forest & $500$ points  & Yes  & 10 &  2.527 \\
DiCE (genetic) & Adult Income & Random Forest & $500$ points  & Yes  & 10 & 2.190 \\
SSBA &  Adult Income  & MLP (2 layers)  & $100$ points  & Yes & 1  & \textbf{2.306} \\
Alibi (Wachter Loss) & Adult Income & MLP (2 layers) & $100$ points  & Yes & 1 & 2.552 \\
\hline
\multicolumn{7}{|c|}{Heart Disease Dataset} \\
\hline
SSBA & Heart Disease & SVM & $100$ points & Yes & 1 & \textbf{18.425} \\
DiCE (kdtree) & Heart Disease & SVM & $100$ points & Yes & 10  & 20.208 \\
DiCE (random) & Heart Disease & SVM & $100$ points & Yes & 10 & 21.654 \\
DiCE (genetic) & Heart Disease & SVM & $100$ points & Yes & 10 & 26.719 \\
SSBA & Heart Disease & MLP (2 layers) & $100$ points  & Yes & 1 & \textbf{6.541} \\
Alibi (Wachter Loss) & Heart Disease & MLP (2 layers) & $100$ points & Yes & 1 & 11.759 \\
\hline
\multicolumn{7}{|c|}{Synthetic Dataset with $20$ features} \\
\hline
SSBA & Synthetic & LR & $500$ points & No & 1 & \textbf{5.404} \\
DiCE (kdtree) & Synthetic & LR & $500$ points & No & 5  & 10.590 \\
DiCE (random) & Synthetic & LR & $500$ points & No & 5 & 10.163 \\
DiCE (genetic) & Synthetic & LR & $500$ points & No & 5 & 10.957 \\
SSBA & Synthetic & MLP (2 layers) & $100$ points  & No & 1 & 2.195 \\
Alibi (Wachter Loss) & Synthetic & MLP (2 layers) & $100$ points & No & 1 & \textbf{0.890} \\
\hline
\end{tabular}
\end{minipage}
\caption{Side-by-side comparison of (a) Computational cost comparison for counterfactual generation, and (b) Euclidean distances of generated counterfactuals on different datasets.}
\end{table*}

\begin{figure*}[ht!]
    \centering
    \begin{subfigure}[b]{0.5\textwidth}
        \centering
        \includegraphics[width=3.5in, height=2.5in]{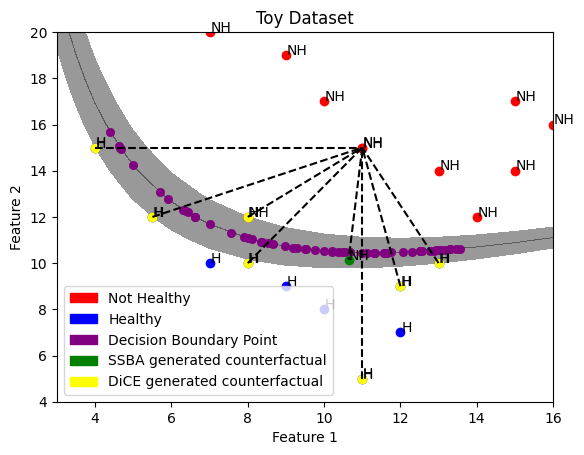}
        \captionsetup{margin=1cm}
        \caption{Comparison of SSBA and DiCE KDTree algorithm (model-agnostic) for the point (11,15) in our toy dataset. }
    \end{subfigure}%
    ~
    \begin{subfigure}[b]{0.5\textwidth}
        \centering
        \includegraphics[width=3.5in, height=2.5in]{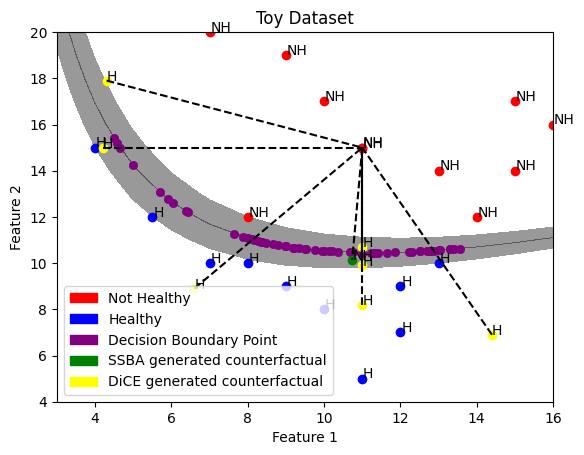}
        \captionsetup{margin=1cm}
        \caption{Comparison of SSBA and DiCE random algorithm (model-agnostic) for the point (11,15) in our toy dataset.}
    \end{subfigure}
    \newline 
    \begin{subfigure}[b]{0.4\textwidth}
        \centering
        \includegraphics[width=3.5in, height=2.5in]{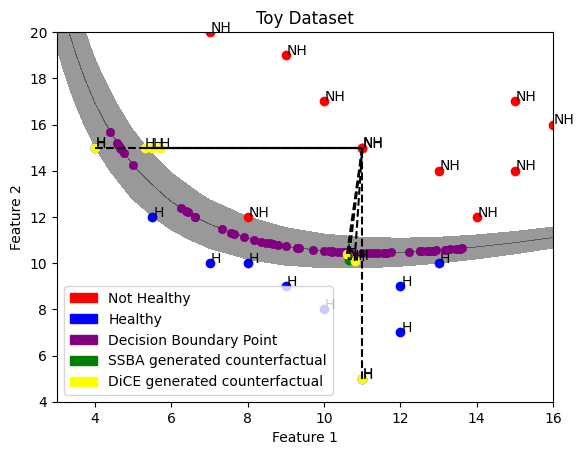}
        \caption{Comparison of SSBA and DiCE genetic algorithm (model-agnostic) for the point (11,15) in our toy dataset. }
    \end{subfigure}%
    \caption{Side-by-side comparison of SSBA with DiCE's model-agnostic approaches. We compare the green point (applying our SSBA method) with the yellow points (applying DiCE's model-agnostic methods). The black decision boundary is generated with matplotlib, and the purple dots are generated with SSBA.}
\end{figure*}

\begin{figure*}[t!]
    \begin{subfigure}[b]{0.5\textwidth}
        \centering
        \includegraphics[width=3.5in, height=2.5in]{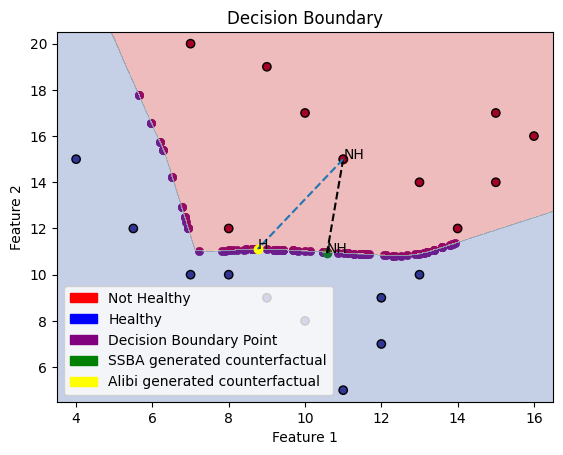}
        \captionsetup{margin=1cm}
        \caption{Comparison of SSBA and Alibi's implementation of Wachter et. al. (2017) objective loss function (gradient-based method) for the point (11,15) in our toy dataset. }
    \end{subfigure}%
    ~
    \begin{subfigure}[b]{0.5\textwidth}
        \centering
        \includegraphics[width=3.5in, height=2.5in]{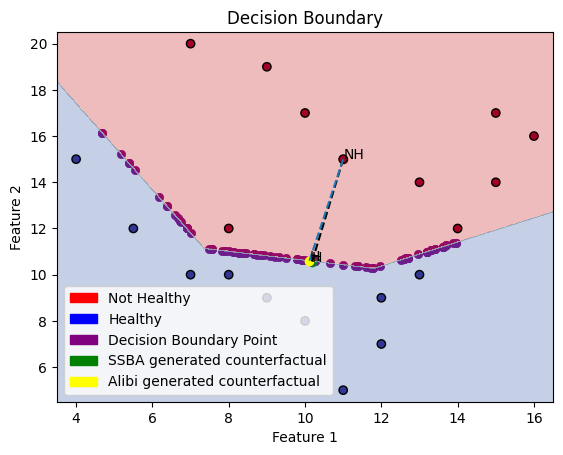}
        \captionsetup{margin=1cm}
        \caption{Comparison of SSBA and Alibi's implementation of Wachter et. al. (2017) objective loss function (gradient-based method) for the point (11,15) in our toy dataset.}
    \end{subfigure}
   \caption{Side-by-side comparison of SSBA with Alibi's gradient approach. We compare the green point (applying our SSBA method) with the yellow point (applying Alibi's gradient method). The black decision boundary is generated with matplotlib, and the purple dots are generated with SSBA.}
\end{figure*} 

\begin{figure*}[t!]
    \begin{subfigure}[b]{0.5\textwidth}
        \centering
        \includegraphics[width=3.5in, height=2.5in]{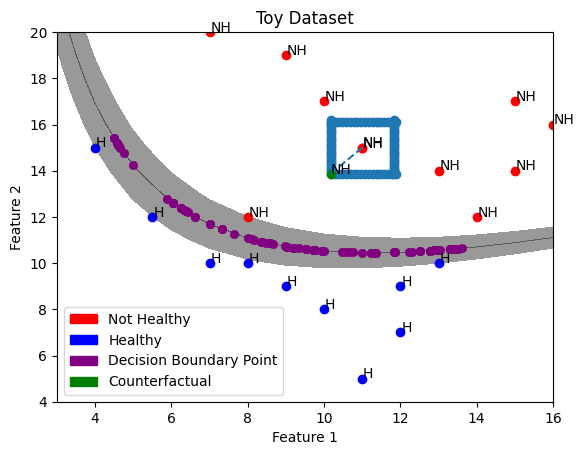}
        \captionsetup{margin=1cm}
        \caption{ In this case, we define $15\%$ constraints in both $x_1$ and $x_2$ directions for the point $(x_1, x_2) = (11,15)$.}
    \end{subfigure}%
    ~
    \begin{subfigure}[b]{0.5\textwidth}
        \centering
        \includegraphics[width=3.5in, height=2.5in]{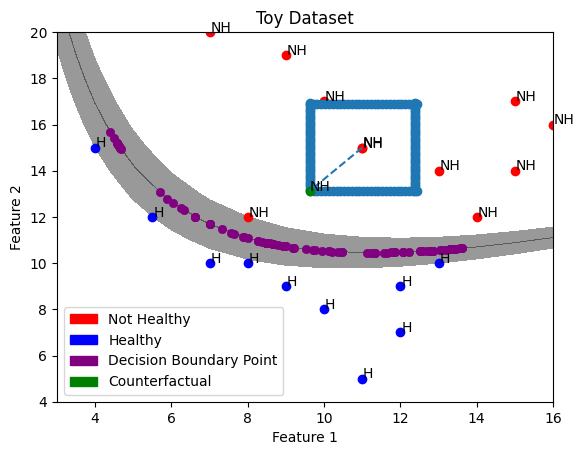}
        \captionsetup{margin=1cm}
        \caption{In this case, we define $25\%$ constraints in both $x_1$ and $x_2$ directions for the point $(x_1, x_2) = (11,15)$.}
    \end{subfigure}
    \caption{Using a trained SVM classifier, we can observe the green point's distance to the boundary. The matplotlib generated decision boundary is colored black. Similar as above, the purple dots are generated with SSBA.}
\end{figure*}

\clearpage
\cleardoublepage
\section{Conclusion}
In the current line of research for counterfactual explanations, much of the recent research in the space of counterfactual generation has focused on optimization (hyperparameter tuning, gradient descent) and the method of tuning the parameters of a loss function that reduces the $L_2$ norm between the original instance and newly generated counterfactual $\mathbf{x'}$ where $n$ is the number of features. In DiCE's case, they have both model-agnostic and gradient-based approaches that use a loss function to optimize the original counterfactual explanation. Their genetic algorithm, one of their three model-agnostic approaches, applies a loss function for optimizing an original set of counterfactuals for gradual improvement across multiple iterations \cite{mothilal2020dice}. Mahajan et. al. [2019] applied both gradient descent and  hyperparameter tuning. They searched for possible counterfactuals using ``random search for $100$ iterations" to best fit the objective that they designed \cite{mahajan2019preserving}. With our work, we propose a model-agnostic method that produces CFEs in high-dimensional space based on the selection of decision boundary points that minimize the $L_2$ norm between the original instance and the boundary point. This method performs well compared to other current model-agnostic methods and optimization techniques with enough boundary points when focusing on the Euclidean metric.

 Against current methods, our approach performs well across datasets of varying dimensionality, and it performs well with added constraints and with no constraints. In the constrained case, our method outperforms DiCE's model-agnostic methods and Alibi's gradient-based method. In the unconstrained case, we underperform relative to Alibi while exceeding DiCE's modeol-agnostic approaches. When compared to DiCE's model-agnostic methods, their methods generate counterfactuals independent of the decision boundary such as generating the counterfactuals using $k$ closest data points (KDTrees algorithm) or generating random counterfactuals (random algorithm).  Our method makes use of information on the boundary in the form of discrete boundary points to generate the nearest CFE. DiCE's method of counterfactual generation does not take into account the boundary which makes it harder to optimize for a minimally distant counterfactual. By comparing Alibi's implementation of Wachter's loss function with our method, we can see that Alibi relies on gradient descent to optimize over a distance-based loss function. While it can generate a reliable CFE, it relies on the fact that the model must be learnable with gradient-based methods. Given that we only generated $10,000$ decision boundary points for each data point compared to Alibi's method, the sparsity of decision boundary points can cause our method to overlook the optimum for the distance. A higher number of decision boundary points can decrease the overall sparsity and find closer points to the boundary. In the bounded case, our method makes use of boundary points to project a point closer to the boundary compared to DiCE and Alibi. While our method underperforms Alibi in the constrained case, our method can build greater robustness with generating a much greater number of points and with higher-dimensional interpolation methods.

While our method does underperform in computational costs, the SSBA originally ran with CPU only, which usually meant that a single run took minutes to complete because it applied binary search to a single point rather than multiple points. However, with recent advancements in GPU support for Scikit-Learn models, we successfully enabled cuML for training the model and generating decision boundary points. This has significantly reduced the latency for boundary point generation by $25-50$ times, allowing it to run in seconds instead of minutes. Although the method produces results at a slower rate than DiCE's model-agnostic methods, the model achieves general improvements in minimizing the distance between the counterfactual and the original instance. 

For finding a counterfactual that matches constraints for feasibility, we take a more straightforward approach that first focuses on the reduction of decision boundary points. Once we generated a set of these points, we applied the constraints and removed those points that do not fit the criteria or the inequalities that we have enforced on the problem. If such a point remains, we return it as a feasible counterfactual explanation. However, in cases where no such feasible points are found, we modify only the continuous features, subject to constraints on the percentage of change within these features. Rather than generating a counterfactual that does not follow feasible constraints, we allow for changes in the continuous features that move towards the decision boundary. As shown in Figure $5$, we generate a ``bounded" counterfactual that moves the query instance slightly towards the boundary. Similarly, DiCE and Alibi also allow users to define bounds for their features.  Mahajan et. al. [2019] also added the possibility of using inequalities as constraints. However without finding points on the boundary, the challenge of finding a bounded counterfactual that would move the original query point closer to the boundary would be harder.

In its current form, our SSBA can outperform prevalent model-agnostic methods in nearest feasible counterfactual explanations and outperforms both DiCE and Alibi's methods for generating ``bounded" counterfactuals. For a synthetic dataset where the labels do not assign meaning to the individual in a dataset, bounded CFEs do not provide much information. However, for our future work, this would be very informative since we intend to use this method to provide a path for intervention. For instance, given a trained model that can diagnose a known disease from observational data, we can use the values of the ``bounded" counterfactuals as a way to direct the individual towards the opposite class, or in this case, a person with no disease. 

With greater knowledge of the decision boundary, our SSBA method can assist in formulating an intervention plan where we identify which features to adjust for directional improvement. 

Our method provides a simple, state-of-the-art model-agnostic method for generating a nearest counterfactual explanation that satisfies required constraints and provides a concrete way to identify directional improvement with the construction of discrete decision boundary points.

\section{Future Works}

When compared to gradient-based methods, the SSBA method can struggle due to sparsity among points in our decision boundary set. However, further extensions of the method, including the use of multi-GPUs, storage and retrieval, and constructing epsilon-balls, can make the method succeed with much more comparable results to gradient-based or hyperparameter search methods while avoiding sparsity among decision boundary points.
 
There are several ways to enhance the SSBA method for more comparable results with optimization methods.

To improve the overall results of the method, we could generate more decision boundary points by creating a topological ``ball" around a given query point in $\mathbf{R}^n$ and doing the same for another point of a different class target. By taking small radii around each point, ensuring that points within the ball belong to two separate classes, we can query and perform binary search on points that are slightly different from the original query points. This is possible since the space $\mathbf{R}^n$ is Hausdorff on the Euclidean metric. This method would assist in generating many more decision boundary points that would reduce sparsity near the boundary. Adding in this concept can boost the method's overall performance in both constrained and unconstrained cases. Another variation of the idea is to sample according to a multivariate Gaussian $\mathcal{N}(\mu = \mathbf{X}, \Sigma)$ where $\mathbf{X} \in \mathbf{R}^n$ is some data point, and then you sample points on different Gaussians. For this method to work, you likely need a small covariance around each mean vector $\mu = \mathbf{X}$.

We can also apply a small shifting factor to the lower bound of the decision boundary points separated by some epsilon $\epsilon$. If we shift the lower bound of points (closer to the original instance), we can sample and choose the minimum of points from the shifts. 

To improve the runtime of decision boundary point generation, we can utilize a multi-GPU setup instead of relying solely on a single GPU for computation. This could speed up the generation of decision boundary points by processing a greater number of batches at once. Additionally, given that the bottleneck for the method is mainly the generation of decision boundary points, if we store boundary points beforehand, we can reuse the points already found and then use computation to construct new points rather than refinding the same points. This would also make experiments comparing our method with other methods much faster since we do not recompute the same boundary points. 

Finally, regarding questions of feasibility and the search for a personalized treatment plan, a method where we can identify mutable features from immutable ones could also provide directional improvement towards a ``no disease" scenario. We can also utilize more inequalities as constraints rather than equalities and use similarity rather than an exact match. Given that finding a decision boundary point with the same categorical features will be unlikely (i.e., finding a person with the same age, same gender, same sex, etc.), we can use inequalities to help find a decision boundary point that is most similar to the original query point. To explain this in simpler terms, if an observational dataset consisted of persons with a known target label for a given disease diagnosis, we might not find a boundary point that equates in all categorical features of the original query (person), such as age, sex, gender, and number of times the person smoked each week. However, if we relax the problem with inequalities, we might find a similar boundary point that provides a case for directional improvement. In a similar sense, Mahajan et. al. [2019] used an inequality for the age of a person in the Adult Income dataset rather than a strict equality \cite{mahajan2019preserving}. 

With the construction of a personalized treatment plan, we intend to use this in combination with an LLM to provide treatment options that shifts the user towards improved health. The SSBA method can be used as a tool for an LLM agent to describe a set of actions or interventions that the user can take to change the features needed for a ``flip" in the diagnosis or model prediction. 

Overall, we build our method upon to provide comparable performance to optimization-based approaches, and with the combination of an LLM to provide actions for the user to take for feature changes, our method can provide a set of ideas of appropriate intervention.

\clearpage 
\newpage
\bibliographystyle{icml2025}
\bibliography{references}

\section{Appendix}

\subsection{Evaluation Metrics} 

In the results section, we share the final results from evaluating various methodologies, including our \textit{Segmented Sampling from Boundary Approximation} approach (SSBA), Alibi's gradient-based approach, and DiCE's model-agnostic approaches. In this section, we present the design of the metrics used to evaluate our method in comparison to Alibi and DiCE's approaches. \\

As shown in Table $1(b)$, we compute the distance column (last column) as an average over all randomly sampled points shown under the number of samples column. 

In the unconstrained case, we compute the scalar distance as an average over the sampled points.

The formula that we use for the scalar in the last column (no constraints) is the one below:
\begin{equation*}
\begin{aligned}
\frac{1}{|B|}*\Sigma_{(x,x') \ \in \ B \ \subseteq \ X_1 \ \times \ X' \ } \ l_2(x, x')
\end{aligned}
\end{equation*}

This dataset $B$ is the sample of points from $X_1 \ \times X'$, where $X_1$ is our set of points that has the label $y = 1$ in our dataset, and $X'$ is the set of all possible generated counterfactuals. In simple terms, we compute the average distances over all data points that are labelled $1$ or ``unhealthy" in the case of the toy dataset. We sample all points from $x \in X_1$ according to a uniform distribution, which means that $x \sim Unif(X_1)$. $x' \in X'$ is the generated counterfactual constructed with $x \in X_1$. For computing the distance, we use $l_2$, which is the Euclidean metric on $\mathbf{R}^{n}$. The $l_2$ norm function computes the Euclidean distance between $x, x' \in \mathbf{R}^n$, where $n$ is the dimension of the space.

In the constrained case, we compare the distance from a given generated counterfactual $x' \in \Delta X'$ to one of the decision boundary points generated with the SSBA method. We define all decision boundary points as points $d \in D$ where $D$ is our total set of generated decision boundary points. $\Delta X'$ is the set of constrained counterfactuals where only the continuous features are constrained and categorical features are left unchanged.

For each of the constrained counterfactuals $ x' \in \Delta X' \subset \mathbf{R}^n$, we compute the distance between each of the decision boundary points $d \in D$ and the counterfactual $x'$. We then select a decision boundary point $d^* \in D$ where $d^* = min_{d \in D} \ l_2 (d, x')$. Once we have found the closest decision boundary point, we then compute the Euclidean distance between that minimally distant point and the generated counterfactual $x' \in \Delta X'$. $B$ represents pairs of points in $D \ \times \ \Delta X'$ where $d^*$ is a minimally distant decision boundary point from a given paired counterfactual $x'$. 

\begin{equation*}
\begin{aligned}
\frac{1}{|B|}*\Sigma_{(d^*,x') \ \in \ B \ \subseteq \ D \ \times \ \Delta X' \ } \ l_2(d, x')
\end{aligned}
\end{equation*}

\subsection{Metric Examples}

\begin{figure}[H]
        \centering
        \includegraphics[width=3in, height=2.5in]{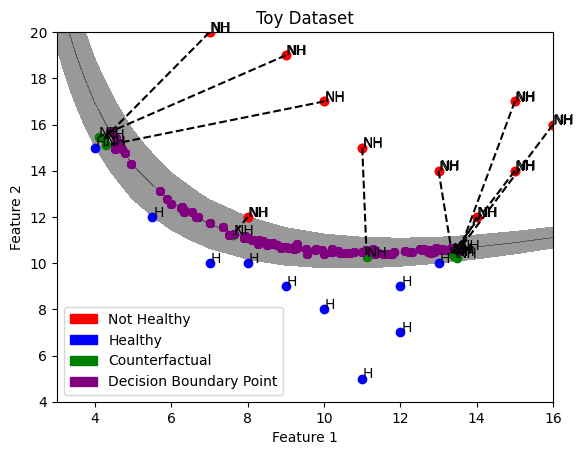}
        \caption{This is an example of how we compute the distance values in the last column of Table $1(b)$ for the unconstrained case. We sample the distances in black where the points are labelled $y = 1$ (or unhealthy in this case). We then average over these distances.}
\end{figure}

 \begin{figure}[H]
        \centering
        \includegraphics[width=3in, height=2.5in]{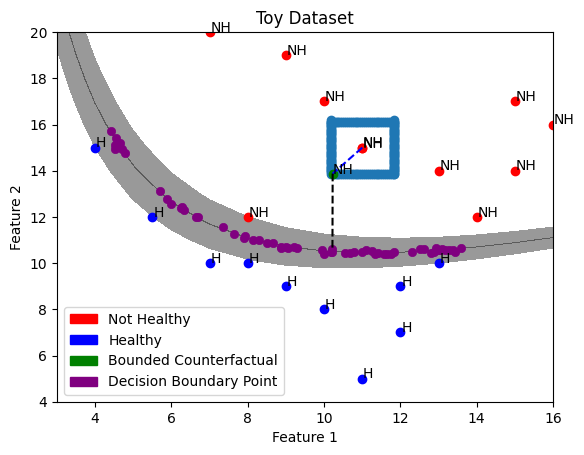}
        \caption{This is an example of how we compute the distance values in the last column of Table $1(b)$ for the constrained case. We sample the distances from the bounded counterfactual highlighted in green to the closest decision boundary point. We sample from the points labelled $y = 1$ (or unhealthy in this case) and average the distances.}
\end{figure}

\subsection{Outline of Proof for Bisection Method Search}\footnote{Vlassopoulos et. al. (2020) do not provide a proof for the bisection method search that they suggested, so we will provide a proof here.}

\textbf{Assumption:  For this proof, we apply the following definition for a decision boundary point $d$. For a point $d$, we define that the $l_2$ distance between $l_2(v_1,d)$ and $l_2(v_2,d)$ as part of two different classes which are both correctly classified and labeled, namely $v_1 \in X_{c_1}^{correct} \subset \mathbf{R}^n$ and $v_2 \in X_{c_2}^{correct} \subset \mathbf{R}^n$, to be less than an sufficient epsilon $\epsilon^*$.
 We must also have a trained model $f$ which generates a decision boundary.
} \\
\textbf{Claim: For a binary classification problem, the bisection method generates an nonempty set $D$ of discrete decision boundary points.} \\
\begin{proof}
The proof for the bisection method search algorithm is based on graph coloring, where we maintain a 2-coloring of a graph $G$ and reduce the $l_2$ distance among points of different classes using binary search. 

Consider a 2-colored graph $G$ from the classes of points that were both correctly classified and labeled $X_{c_1}^{correct}$ and $X_{c_2}^{correct}$. Let $X_{c_1}^{correct}$ be the set of nodes whose coloring is red, and let $X_{c_2}^{correct}$ be the set of nodes whose coloring is blue. Let the edges of the graph $G$ be between nodes of different colors, and let every edge $(v_1, v_2) \in X_{c_1}^{correct} \times X_{c_2}^{correct}$ be an edge in the graph $G$.

Then consider an edge $(v_1, v_2)$ between two vertices $v_1 \in X_{c_1}^{correct}$ and $v_2 \in X_{c_2}^{correct}$. Take the midpoint of these two vertices, and we find a new point $\frac{v_1 + v_2}{2} = x$ for which we do not know the label. With the trained model $f$, we apply the model to the point $x$ such that $f$ produces a label $f(x)$ of one of two colors, red or blue. 

In the case that the label for $x$ is red, we replace the endpoint $v_1$ that was originally colored red with our new midpoint $x$ such that our new smaller edge becomes $(x, v_2)$. This new edge maintains the coloring, and the new $l_2$ norm is now $l_2(x,v_2) < l_2(v_1, v_2)$. 

In the case that the label for $x$ is blue, we replace the endpoint $v_2$ that was originally blue with our new midpoint $x$ such that we have a smaller edge $l_2(v_1, x) < l_2(v_1, v_2)$ while maintaining the coloring of $G$.

Without loss of generality, we can repeat this process for new midpoints such that for step $k$, a given edge $(v_1^{k+1}, v_2^{k+1})$ will have half the norm of $(v_1^{k}, v_2^{k})$. Because of this repeated halving of the interval, we know that $l_2(v_1^{k}, v_2^{k}) = \frac{l_2(v_1, v_2)}{2^k}$. We can choose $N$ such that $\frac{l_2(v_1^{1}, v_2^{1})}{2^N} < \epsilon^*$. By algebra, we can rearrange to find that $log_2 \frac{l_2(v_1, v_2)}{\epsilon^*} < N$. If we choose $N^* = \lfloor log_2 \frac{l_2(v_1, v_2)}{\epsilon^*} \rfloor + 1$ such that $N^* \geq N$, we have found an integer $N^*$ that ensures the two endpoints of the edge are separated by an $l_2$ norm less than $\epsilon^*$. After $N^*$ iterations, we take the midpoint $M$ of $v_1^{N^*}$ and $v_2^{N^*}$ such that $l_2(M, v_1^{N*}) < \epsilon^*$ and $l_2(M, v_2^{N*}) < \epsilon^*$. Thus, $M$ represents a decision boundary point because $v_1^{N*}$ and $v_2^{N*}$ are points of different colors separated by less than $\epsilon^*$. Doing this for all edges in $G$, we get a discrete set of boundary points $D$.
\end{proof}

\newpage 
\subsection{Additional Properties of SSBA}

\subsubsection{Reasonable Feature Bounds on Decision Boundary Points}
\noindent \\
\textbf{Assumptions: Same as above.} \\
\\
\textbf{Claim: For a dataset $X$ where $X_i$ refers to all observations of a given feature $i$, we can guarantee that the SSBA method generates decision boundary points in a ``reasonable'' range, which we define to be}

\[
d_i \in [\min(X_i), \max(X_i)] \quad \forall i \in \{1, \dots, n\},
\] 
for a decision boundary point 
\[
d = (d_1, d_2, \dots, d_n), \ d_i \in \mathbb{R}, \ d \in \mathbb{R}^n.
\]

\begin{proof}
Let $\min(X_i)$ and $\max(X_i)$ be the minimum and maximum values of feature $i$ across the dataset, for each $i \in \{1, \dots, n\}$.  \\

Take a decision boundary point $d = (d_1, \cdots,d_n)$ generated with the SSBA approach. By construction, after a sufficient number of iterations, we have
\[
d_i = \alpha x_i + (1-\alpha)y_i, \quad \text{for some } \alpha \in [0,1],
\]
where $x_i, y_i \in X_i$.  \\
\\

\noindent 
Since $x_i, y_i \in [\min(X_i), \max(X_i)]$, it follows that
\[
\min(X_i) \leq x_i \leq \max(X_i) \quad \text{and} \quad \min(X_i) \leq y_i \leq \max(X_i).
\]  \\
\noindent
Therefore,
\[
d_i = \alpha x_i + (1-\alpha)y_i \leq \alpha \max(X_i) + (1 - \alpha) \max(X_i) 
= \max(X_i),
\] 
\[
d_i = \alpha x_i + (1-\alpha)y_i \geq \alpha \min(X_i) + (1-\alpha)\min(X_i)
= \min(X_i).
\] 
\noindent \\
Thus,
\[
\min(X_i) \leq d_i \leq \max(X_i).
\] \\
\noindent
Since this holds for all $i \in \{1, \dots, n\}$, we conclude that
\[
d_i \in [\min(X_i), \max(X_i)] \quad \forall i \in \{1, \dots, n\}.
\] 
\end{proof}

\subsubsection{Generating More Decision Boundary Points Than Number of Data Points in the dataset}

Due to the $N^2$ nature of the algorithm, our method is generally capable of generating more boundary points than there are data points in our dataset. We can generally ensure that we can generate a number of data points such that $|D| >> N$ where $|D|$ is the cardinality or the size of our decision boundary set and $N$ is the number of total data points in our original dataset. The only exception to this would be a heavily skewed dataset like 1 dataset in one class and the $N-1$ data points in the other class. Then we would generate only $N-1$ boundary points. But, for a balanced dataset, we will have this property. This becomes especially true for balanced datasets with large $N$ such that $N^2$ nature for the algorithm takes over and enables a much greater generation of discrete boundary points.

\end{document}